\documentclass[conference]{IEEEtran}
\usepackage{times}

\usepackage{cite}
\usepackage{multicol}
\usepackage[bookmarks=true]{hyperref}
\usepackage{graphicx}
\usepackage{algorithm}
\usepackage{algpseudocode}
\usepackage{amssymb}
\usepackage{amsmath}
\usepackage{xcolor}
\usepackage{subcaption}
\newcommand{\algoname}{PSANE}

\usepackage{eso-pic}
\AddToShipoutPictureBG*{%
  \AtPageLowerLeft{%
    \put(\LenToUnit{0.75in},\LenToUnit{0.4in}){%
      \parbox[b]{\dimexpr\paperwidth-1.5in\relax}{\centering\scriptsize
        Accepted to the 2026 IEEE/RSJ International Conference on Intelligent Robots and Systems (IROS).\\
        \textcopyright~2026 IEEE. Personal use of this material is permitted. Permission from IEEE must be obtained for all other uses, in any current or future media, including reprinting/republishing this material for advertising or promotional purposes, creating new collective works, for resale or redistribution to servers or lists, or reuse of any copyrighted component of this work in other works.}}}}
\begin{document}

\title{Proprioceptive Safe Active Navigation and Exploration for Planetary Environments}

\author{
\IEEEauthorblockN{
Matthew Y. Jiang\IEEEauthorrefmark{1},
Feifei Qian\IEEEauthorrefmark{2},
Shipeng Liu\IEEEauthorrefmark{2}\IEEEauthorrefmark{4}
}
\IEEEauthorblockA{
\IEEEauthorrefmark{1}Georgia Institute of Technology, Atlanta, GA 30332, USA
}
\IEEEauthorblockA{
\IEEEauthorrefmark{2}University of Southern California, Los Angeles, CA 90089, USA
}
\IEEEauthorblockA{
\IEEEauthorrefmark{4}Corresponding author
}
\IEEEauthorblockA{
Email: matthewjiang@gatech.edu, \{shipengl, feifeiqi\}@usc.edu
}
}
\maketitle

\begin{abstract}
Deformable granular terrains introduce significant locomotion and immobilization risks in planetary exploration and are difficult to detect via remote sensing (e.g., vision). Legged robots can sense terrain properties through leg--terrain interactions during locomotion, offering a direct means to assess traversability in deformable environments. How to systematically exploit this interaction-derived information for navigation planning, however, remains underexplored. We address this gap by presenting \algoname, a \textit{P}roprioceptive \textit{S}afe \textit{A}ctive \textit{N}avigation and \textit{E}xploration framework that leverages leg--terrain interaction measurements for safe navigation and exploration in unknown deformable environments. \algoname{} learns a traversability model via Gaussian Process regression to estimate and certify safe regions and identify exploration frontiers online, and integrates these estimates with a reactive controller for real-time navigation. Frontier selection is formulated as a multi-objective optimization that balances safe-set expansion probability and goal-directed cost, with subgoals selected via scalarization over the Pareto-optimal frontier set. \algoname{} safely explores unknown granular terrain and reaches specified goals using only proprioceptively estimated traversability, while achieving performance improvements over baseline methods.
\end{abstract}

\IEEEpeerreviewmaketitle
\section{Introduction}
\label{sec:Introduction}

Planetary exploration systems rely heavily on exteroceptive sensing modalities such as vision and LiDAR for mapping and safe traversal. For example, the Mars 2020 rover employs HazCams and NavCams for geometric reconstruction and hazard detection \cite{maki2020mars2020}. More broadly, robotic navigation frameworks typically construct geometric maps using visual--inertial SLAM or LiDAR pipelines \cite{mur2015orb, campos2021orb}, and infer traversability from geometric features such as slope, roughness, step height, and local surface structure \cite{wermelinger2016navigation, wang2023towards, fahmi2022vital, wellhausen2020safe}, sometimes augmented with semantic or material recognition cues \cite{guan2022ga}. These approaches have been effective in rigid environments where mobility constraints are strongly correlated with observable surface geometry.

On deformable terrains, however, surface geometry and appearance are often insufficient for estimating traversal risks. Locomotion performance in granular soils and regolith-like media depends on subsurface mechanical properties such as shear strength, compaction state, and sinkage resistance \cite{zhuang2022review, liu2025bio, liu2023mud}. These properties are not reliably inferable from remote sensing alone \cite{fulcher2025effect}. The Spirit rover's entrapment \cite{callas2015spirit} exemplifies this limitation: visually benign terrain can still induce immobilization due to weak underlying layers. These failures motivate moving beyond geometry-only risk assessment.

\begin{figure}[h]
    \centering
    \includegraphics[width=0.85\linewidth]{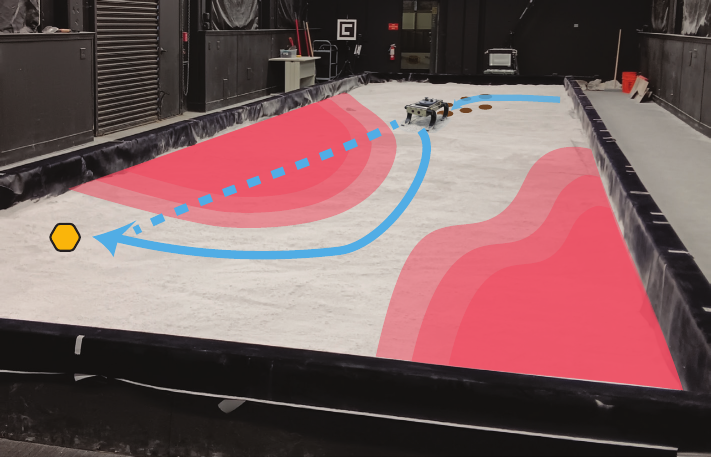}
    \caption{Overview of safe reactive navigation.
(Red) Unsafe zones for the robot to avoid.
(Brown) Measured terrain penetration resistance.
(Blue Solid) Executed and planned paths.
(Blue Dotted) Naive path to goal.
(Yellow) Navigation Goal.}
    \label{fig:overview}
\end{figure}

Recent work has explored interaction-informed perception by leveraging proprioceptive signals, including motor positions and currents, to either classify terrain types \cite{dey2022prepare, fu2022coupling} or estimate terrain mechanical properties through continuous spatial inference \cite{fulcher2025effect, 2024LPICo3040.1806F, 2023LPICo2806.1733B, elnoor2024pronav, qian2019rapid, 10114536109773635112, 1011453623383} based on robot–ground interaction dynamics. In existing work, interaction-derived information is primarily used for locomotion adaptation or low-level control rather than navigation-level decision-making. Even when interaction-derived estimates inform higher-level planning in multi-robot settings~\cite{liu2026scout}, the scouting robot’s own navigation problem, which must balance exploration and safety under uncertainty, remains largely underexplored.


This paper considers navigation in settings where proprioception is the primary source of terrain information. Because contact measurements are sparse, local, and available only at discrete contact locations, the robot must transform them into a spatial risk representation, certify traversable regions, actively collect informative data, and progress toward the goal while remaining safe. Gaussian Processes (GPs) support this uncertainty-aware inference from sparse interaction data\cite{leininger2024gaussianprocessbasedtraversabilityanalysis, ghaffari2018gaussian}. Prior GP-based safe learning mainly selects safe query points that expand knowledge while avoiding unsafe regions \cite{safeopt, 10.1007/978-3-319-23461-8_9, wachi2018safe, 9767193}; deformable-terrain navigation instead requires synthesizing safe, dynamically feasible trajectories that expand the certified safe set while moving toward a spatial goal.
To this end, we present \algoname, a \textit{P}roprioceptive \textit{S}afe \textit{A}ctive \textit{N}avigation and \textit{E}xploration framework for safe navigation and exploration in unknown deformable planetary environments using only interaction-derived information. \algoname{} learns a traversability model online via GP regression from proprioceptive leg--terrain measurements, and uses the resulting predictive mean and uncertainty to conservatively certify safe regions and identify frontier regions for exploration. A frontier-based subgoal selection strategy balances goal-directed progress with safe-region expansion, and is integrated with a diffeomorphic reactive controller \cite{arslan2016sensorbasedreactive, vasilopoulos2021reactivenavigationpartiallyfamiliar} that outputs real-time velocity commands while treating uncertified regions as obstacles.

Our main contributions are:
\begin{itemize}
\item A closed-loop formulation of safe navigation on deformable terrain that integrates proprioceptive risk estimation, uncertainty-aware safe-set certification, active frontier selection, and reactive execution.
\item A multi-objective frontier subgoal selection strategy that balances safe-set expansion and goal progress, integrated with a reactive controller for real-time safe navigation.
\item Extensive simulation validation using field-measured LHS-1 regolith terramechanics data, demonstrating reliable identification of high-risk regions and safe exploration of unknown environments.
\end{itemize}

\begin{figure*}[t]
    \centering
    \includegraphics[width=0.99\linewidth]{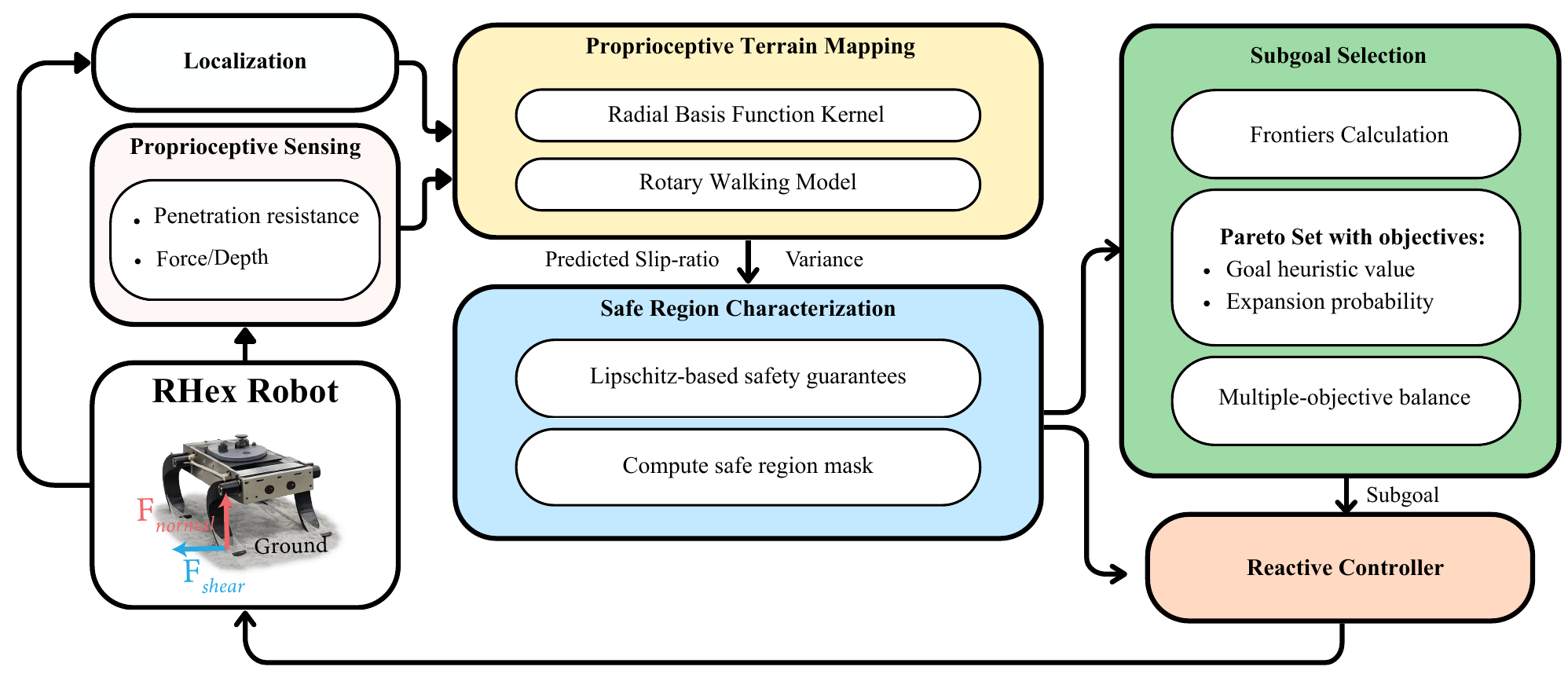}
    \caption{Overview of the proposed PSANE navigation framework.
}

    \label{fig:overview_pipeline}
\end{figure*}

\begin{figure*}[t]
    \centering
    \includegraphics[width=0.89\linewidth]{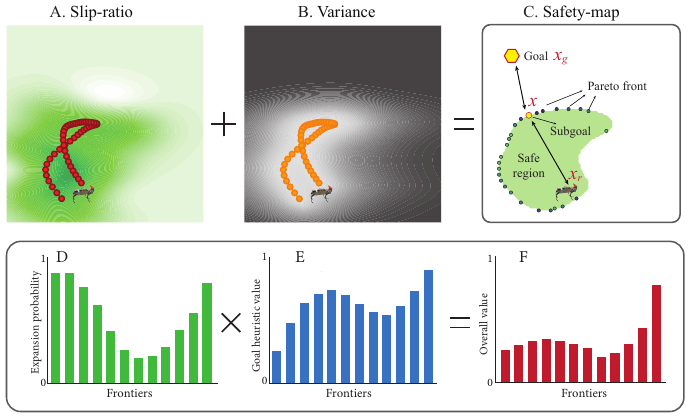}
    \caption{Terrain-aware risk estimation and safety-aware planning. A GP infers (A) slip ratio and (B) uncertainty. (C) Confidence bounds define a conservative safety map and Pareto frontier set. (D--F) Frontier expansion probability and goal heuristic are combined into the overall score.}

    \label{fig:pipeline_detailed}
\end{figure*}
\section{Problem Formulation}
\label{sec:problem_formulation}
We consider a legged robot~\cite{saranli2001rhex} operating in an unknown deformable granular terrain with spatially varying strength (Fig.~\ref{fig:overview}). In such environments, locomotion performance depends strongly on local substrate mechanics, particularly terrain penetration resistance~\cite{li2009sensitive}. During locomotion, the robot interacts intermittently with the terrain through discrete leg--ground contacts. At each contact event, it can estimate the local terrain penetration resistance from proprioceptive force--depth measurements. The estimated penetration resistance is then used to predict locomotion performance, including forward speed and slip ratio $s$, using an existing ``rotary-walking'' model~\cite{li2009sensitive}, validated on dry granular media across compaction levels; other substrate regimes may require re-identifying its parameters. Since large slip ratios indicate degraded traction and possible immobilization, regions that induce high slip are considered high-risk.

Based on this discrete sensing model, we formulate safe navigation as a probabilistic learning and control problem over a bounded planar workspace $\mathcal{W} \subset \mathbb{R}^2$.
For computation, we discretize the workspace into a finite set of query points
$D \subset \mathcal{W}$, and perform all GP inference and safe-set operations on $D$. Let $f : \mathcal{W} \rightarrow \mathbb{R}$ denote an unknown slip-ratio field, where larger $f(\mathbf{x})$ corresponds to larger slip and thus higher immobilization risk. The robot can estimate $f(\mathbf{x})$ only at contacted locations $\mathbf{x}_t$ through noisy measurements:
\begin{equation}
\label{eq:measurements}
y_t = f(\mathbf{x}_t) + \varepsilon_t, 
\quad \varepsilon_t \sim \mathcal{N}(0, \sigma_{\text{noise}}^2).
\end{equation}

From measurements collected up to time $t$, the robot maintains a probabilistic slip-ratio map
\begin{equation}
\mathcal{T}_t = \left\{\mu_t(\mathbf{x}), \sigma_t(\mathbf{x}) \mid \mathbf{x} \in \mathcal{W}\right\},
\end{equation}
where $\mu_t(\mathbf{x})$ and $\sigma_t(\mathbf{x})$ denote the posterior mean and uncertainty, respectively. Given a safety threshold $h$ ($h=0.5$ in simulation), a location is safe if $f(\mathbf{x}) \leq h$, leaving margin before near-immobilization; smaller $h$ is more conservative but may limit exploration. The objective is to use $\mathcal{T}_t$ to safely expand the certified safe region and navigate to a goal $\mathbf{x}_g \in \mathcal{W}$.

\section{Proprioceptive Safe Navigation and Exploration }
Building on Sec.~\ref{sec:problem_formulation}, \algoname{} consists of (i) GP-based online traversability mapping from proprioceptive leg--terrain interactions, (ii) uncertainty-aware safe-set certification and frontier extraction, and (iii) multi-objective frontier subgoal selection integrated with a reactive controller for real-time navigation (Fig.~\ref{fig:overview_pipeline}). The terrain mapping and slip-ratio estimation components build upon existing interaction-based modeling methods (e.g., \cite{fulcher2025effect}) and are integrated here as a perception module within the broader safe navigation framework.

The mapping module fits force–depth curves to estimate penetration resistance, converts the resulting strength map into slip-ratio risk, and uses uncertainty to certify safe regions and generate frontier subgoals. Each frontier is scored by safe-set expansion probability and normalized goal cost; the selected subgoal is passed to a diffeomorphic reactive controller~\cite{arslan2016sensorbasedreactive, vasilopoulos2021reactivenavigationpartiallyfamiliar}. While terrain-strength estimation and the rotary-walking slip model follow established formulations~\cite{li2009sensitive}, this work uses their uncertainty-aware risk estimates for conservative safe-set certification and multi-objective frontier navigation.

\subsection{Proprioceptive Terrain Mapping}
\label{sec:ProprioceptiveMapping}
Given the measurement history $\mathcal{D}_t = \{(\mathbf{x}_i, y_i)\}_{i=1}^t$ from Equation~\ref{eq:measurements}, the terrain penetration resistance and slip ratio $\mathcal{T}_t$ are maintained as a Gaussian Process (GP) posterior using the RBF kernel,
$
k(\mathbf{x}, \mathbf{x}') = \sigma_f^2 \exp\left(-\frac{\|\mathbf{x} - \mathbf{x}'\|^2}{2\ell^2}\right)
$
($\ell = 0.6$\,m, $\sigma_f = 0.5$, $\sigma_{\text{noise}} = 0.01$ in our experiments).
The GP provides predictive means $\mu_t(\mathbf{x})$ and variances $\sigma_t^2(\mathbf{x})$ at any location $\mathbf{x}$ for terrain-strength and slip-ratio predictions, which guide safe exploration. (Fig.~\ref{fig:pipeline_detailed}A, Fig.~\ref{fig:pipeline_detailed}B) 
\subsection{Confidence-guided Safe Region Characterization}
\label{sec:ConfidenceGuided}
To characterize which areas are safe for traversal, we first quantify the confidence of the predicted slip ratio at each location $\mathbf{x}$ using the predictive mean $\mu_{t-1}(\mathbf{x})$ and the predictive standard deviation $\sigma_{t-1}(\mathbf{x})$. We then compare this confidence interval with a predefined safety threshold $h$ to determine whether a location can be certified as safe. 
\subsubsection{Slip-ratio Prediction Confidence Interval}
A confidence parameter $\beta > 0$ is introduced to control how conservatively the predicted slip ratio is interpreted. The uncertainty-aware confidence bounds are defined as $c_t(\mathbf{x}) = \left[\mu_{t-1}(\mathbf{x}) \pm \sqrt{\beta}\, \sigma_{t-1}(\mathbf{x})\right]$, and are iteratively intersected with prior confidence sets to ensure consistency, $C_t(\mathbf{x}) = C_{t-1}(\mathbf{x}) \cap c_t(\mathbf{x})$, initialized with $C_0(\mathbf{x}) = (-\infty,\infty)$ for all $\mathbf{x}\in D$.

From $C_t(\mathbf{x})$, the lower and upper bounds of slip ratio are defined as $\ell_t(\mathbf{x}) = \min C_t(\mathbf{x})$ and $u_t(\mathbf{x}) = \max C_t(\mathbf{x})$.

Because higher slip indicates higher traversal risk, $u_t(\mathbf{x})$ is the worst-case slip consistent with model uncertainty, and $\mathbf{x}$ is certified safe only if $u_t(\mathbf{x}) \le h$, meaning that even the largest plausible slip estimate remains below the safety threshold. The lower bound $\ell_t(\mathbf{x})$ gives an optimistic traversability estimate, while the interval width measures uncertainty and highlights regions where additional exploration may reduce risk; larger $\beta$ yields more conservative certification ($\beta=3$ in our experiments).

\subsubsection{Safe Region Characterization}
At iteration $t$, safety is certified conservatively using the GP upper bound $u_t(\mathbf{x})$. Assuming the true slip field is Lipschitz continuous with constant $L$~\cite{safeopt}, a point $\mathbf{x}'$ is certified safe if it can be reached from a previously safe point $\mathbf{x}\in S_{t-1}$ with $u_t(\mathbf{x}) + L\|\mathbf{x}-\mathbf{x}'\| \le h$. This condition propagates safety from certified locations to nearby states using the GP worst-case slip and a local Lipschitz prior at the map resolution. The Lipschitz constant bounds how rapidly slip can vary spatially within a terrain regime, but is not a global smoothness assumption over all terrain types; abrupt regime changes may require conservative bounds or complementary terrain segmentation. We set $L=2$ from the ground-truth slip gradients; in practice, $L$ must be chosen conservatively from prior terrain knowledge, trading slower expansion for a preserved safety guarantee. Environment~2 (Sec.~\ref{sec:Results}) deliberately stresses this assumption with abrupt transitions; the certified methods incurred no safety violations there.

The safe region is therefore updated as
\begin{equation}
S_t = \bigcup_{\mathbf{x} \in S_{t-1}}
\left\{
\mathbf{x}' \in D
\,\middle|\,
u_t(\mathbf{x}) + L \|\mathbf{x}-\mathbf{x}'\| \leq h
\right\}.
\end{equation}

The robot begins with an initial safe set $S_0$ from its starting location and measurements. As observations reduce uncertainty, upper bounds shrink and the safe set expands into previously uncertified areas.

\subsection{Frontiers Detection and Subgoal Selection} 
\label{sec:FrontierDetection}
\subsubsection{Frontiers Detection}
Frontiers are defined as the boundaries between the certified safe region and uncertified terrain, representing locations where safe-set expansion may occur. We apply the Suzuki border-following algorithm~\cite{suzuki1985topological} to the discretized safety map to extract these boundaries as frontier candidates.

Since the safe region may contain multiple disjoint components, we retain only frontier points in the same connected safe component as the robot, denoted by $G_t$, ensuring that selected subgoals are physically reachable within certified safe terrain.
\subsubsection{Subgoal Selection}
\label{sec:FrontierSubgoal}
Ideally, an optimal frontier $\mathbf{x}_t^*$ would minimize the total start-to-goal path cost under safety constraints; in practice it cannot be computed, since the true cost-to-go depends on slip ratios and motion costs beyond the currently certified safe region.

To address this uncertainty, we approximate the long-horizon planning problem using heuristic objectives that balance goal progress and safe-region expansion (Fig. \ref{fig:pipeline_detailed}C). At each planning step, the robot evaluates all frontier candidates $\mathbf{x} \in G_t$ according to two complementary criteria: an expansion probability and a goal heuristic value.

The \textbf{expansion probability}, shown in Fig.~\ref{fig:pipeline_detailed}D, is defined as $P_e(\mathbf{x}) = 1 - \exp(-k_e\, g_t(\mathbf{x}))$, where $k_e > 0$ (0.1 in our implementation) controls the growth rate and ensures $P_e(\mathbf{x}) \in [0,1)$ with diminishing returns as the expansion potential increases. The quantity $g_t(\mathbf{x})$ measures the conservative expansion capability of frontier $\mathbf{x}$ and is defined as
\begin{equation}
g_t(\mathbf{x}) =
\left|
\left\{
\mathbf{x}' \in D \setminus S_t 
\;\middle|\;
\ell_t(\mathbf{x}) + L \|\mathbf{x}-\mathbf{x}'\| \le h
\right\}
\right|,
\end{equation}
which counts the number of previously uncertified states ($D \setminus S_t$) that can be conservatively guaranteed safe by expanding from frontier $\mathbf{x}$. In practice, evaluation is restricted to uncertified states within the Lipschitz-certified radius $r(\mathbf{x}) = (h - u_t(\mathbf{x}))/L$, since states outside this radius cannot satisfy the safety condition. This restriction reduces computational complexity while preserving the conservative safety guarantee.

The \textbf{goal heuristic value function}, shown in Fig.~\ref{fig:pipeline_detailed}E, is defined as
\begin{equation}
V_{\mathrm{goal}}(\mathbf{x}) 
= \exp\!\left(-k_g \left( 
\|\mathbf{x}-\mathbf{x}_g\| 
+ \|\mathbf{x}-\mathbf{x}_r\|
\right)\right),
\end{equation}
where $\mathbf{x}_g$ denotes the mission goal (Fig. \ref{fig:pipeline_detailed}C), $\mathbf{x}_r$ is the current robot position, and $k_g > 0$ (0.1 in our implementation) is a decay parameter controlling the emphasis on shorter paths. The exponential form ensures that $V_{\mathrm{goal}}(\mathbf{x}) \in (0,1]$, yielding a bounded and smoothly decreasing preference for frontier points that are farther from the goal or more costly to reach. When the robot operates in pure exploration mode without a specified mission goal, the distance-to-goal term is removed and the value function reduces to $V_{\mathrm{goal}}(\mathbf{x}) = \exp(-k_g \|\mathbf{x}-\mathbf{x}_r\|)$, which prioritizes frontiers that are locally efficient to reach.

Given the two value functions, we first compute the Pareto front~\cite{ngatchou2005pareto} with respect to 
$P_e(\mathbf{x})$ and $V_{\mathrm{goal}}(\mathbf{x})$, thereby preserving the 
multi-objective structure. From the resulting Pareto set, we evaluate two 
selection strategies. 

The first strategy selects a frontier according to a predefined priority order 
between the objectives (denoted as PGH in Sec.~\ref{sec:Results}). The second strategy scalarizes the objectives by defining an overall frontier score $V_{\mathrm{overall}}(\mathbf{x}) = V_{\mathrm{goal}}(\mathbf{x}) \cdot P_e(\mathbf{x})$ (PSANE), and selects the frontier with the highest $V_{\mathrm{overall}}$.
We compare the performance of these two strategies in Sec.~\ref{sec:Results}.

\begin{algorithm}[h]
\caption{Proprioceptive Safe Active Navigation and
Exploration (\algoname{}).}
\label{alg:srpgt}
\begin{algorithmic}[1]
\Require Workspace $D$, GP prior $(\mu_0,k,\sigma_0)$, Lipschitz $L$, initial safe set $S_0$, safety threshold $h$, goal $\mathbf{x}_g$, decay parameters $k_g,k_e>0$
\State Initialize GP posterior; $S_0\gets S_0$
\For{$t=1,2,\ldots$}
    \State Compute GP bounds $\ell_t,u_t$ on $D$; $S_t\gets\mathrm{SafeExpand}(S_{t-1},u_t,L,h)$
    \State $\mathcal{O}_t\gets D\setminus S_t$; $G_t\gets\mathrm{Frontier}(S_t,\mathbf{x}_r(\tau))$
    \ForAll{$\mathbf{x}\in G_t$}
        \State $V_{\mathrm{goal}}(\mathbf{x})\gets\exp\!\big(-k_g(\|\mathbf{x}-\mathbf{x}_g\|+\|\mathbf{x}-\mathbf{x}_r\|)\big)$
        \State $r(\mathbf{x})\gets\max\!\big(0,\frac{h-\ell_t(\mathbf{x})}{L}\big)$
        \State $g_t(\mathbf{x})\gets |\{\mathbf{x}'\in D\setminus S_t\mid \|\mathbf{x}-\mathbf{x}'\|\le r(\mathbf{x})\}|$
        \State $P_e(\mathbf{x})\gets 1-\exp(-k_e\,g_t(\mathbf{x}))$
        \State $V_{\mathrm{overall}}(\mathbf{x})\gets V_{\mathrm{goal}}(\mathbf{x})\cdot P_e(\mathbf{x})$
    \EndFor
    \State $\mathbf{x}_t\gets\arg\max_{\mathbf{x}\in G_t}V_{\mathrm{overall}}(\mathbf{x})$
    \State Navigate to $\mathbf{x}_t$ using obstacle set $\mathcal{O}_t$
    \State Update $\mathcal{D}_t$ with $(\mathbf{x}_t,y_t)$ ; update GP with $\mathcal{D}_t$
\EndFor
\end{algorithmic}
\end{algorithm}

\subsection{Reactive Navigation}
The planner provides subgoals, while a reactive controller adapted from \cite{vasilopoulos2021reactivenavigationpartiallyfamiliar} generates real-time motion without repeated global replanning.

\subsubsection{Geometric Obstacle Construction}
\begin{figure}
    \centering
    \includegraphics[width=1\linewidth]{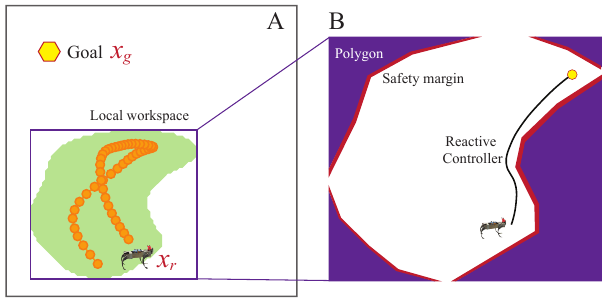}
    \caption{Integration with the reactive controller. The certified safe region (green) is converted to buffered and simplified obstacle polygons (purple), from which the controller generates forward and angular velocity commands.}
    \label{fig:bounds}
\end{figure}

For reactive navigation, the certified safe region $S_t$ is converted into continuous free space (Fig.~\ref{fig:bounds}A). To remain conservative, we shrink the outer boundary, enlarge interior holes by a fixed safety margin, and simplify polygons while preserving topology. Formally, the controller receives $\mathcal{O}_t = \mathcal{W} \setminus S_t$ as obstacles; its collision-avoidance guarantee thus confines closed-loop trajectories to the certified safe set $S_t$ (up to the buffer margin).

\subsubsection{Diffeomorphic Reactive Control}

We adopt the diffeomorphic reactive navigation framework from \cite{vasilopoulos2021reactivenavigationpartiallyfamiliar}, which maps the physical workspace to a simplified model space for potential-field control. For polygonal obstacles, smooth model-space trajectories correspond to obstacle-free physical paths, enabling real-time control as the obstacle set evolves.

\subsection{System Integration}
The complete system integrates GP terrain mapping, frontier subgoal selection, and local reactive control: measurements update the terrain model, the planner selects a certified subgoal, and the controller generates motion commands (Algorithm~\ref{alg:srpgt}).

\section{Results}
\label{sec:Results}
\subsection{Simulation Setup}
\begin{figure}[h]
    \centering
    \includegraphics[width=1\linewidth]{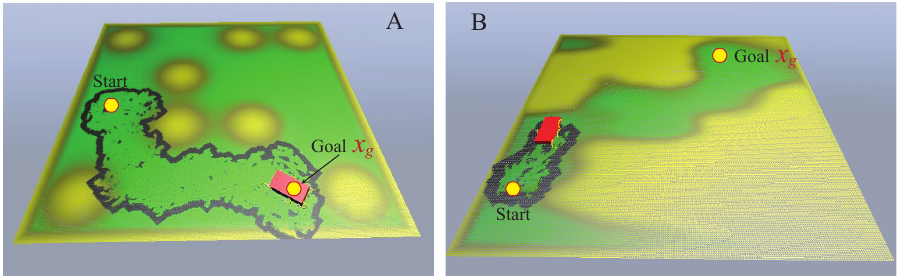}
    \caption{Chrono simulation environments. Environment 1 (A) features smoother terrain variations, whereas Environment 2 (B) exhibits sharper spatial changes in terrain properties. Green denotes high penetration resistance (LHS-1), and yellow denotes low.}
    \label{fig:chrono}
\end{figure}

Terrain penetration resistance is critical for robot performance on highly deformable substrates, such as granular media common in planetary environments. However, preparing a deformable substrate at controlled compaction levels is labor-intensive and difficult to reproduce consistently, limiting large-scale data collection for navigation experiments. To address this challenge, we used a Chrono simulation \cite{10.1007/978-3-319-40361-8_2} (Fig.~\ref{fig:chrono}) which supports multiple terramechanics models. 
To emulate lunar simulant behavior under different compaction levels, we parameterize the Chrono terrain model using experimentally measured regolith mechanical properties reported in~\cite{liu2026scout}. Specifically, we instantiate a Soil Contact Model (SCM) terrain in Chrono~\cite{serban2023real, batagoda2025physics}, which supports spatially varying terrain parameters over a 2D domain. We then simulate legged locomotion on the validated SCM terrain. Chrono computes the interaction forces between the robot’s legs and body and the deformable terrain, and integrates the full-body dynamics to evolve both the robot and terrain states.

\subsection{Experimental Design}
In this section, we evaluate the performance and efficiency of \algoname\ against three baselines: the Naive Goal Heuristic (NGH), the Safe-Goal Heuristic (SGH), and the Pareto-Goal Heuristic (PGH) for goal-directed navigation tasks. For safe terrain exploration without a predefined goal, we compare \algoname\ against PGH, which selects the frontier with maximum $P_e(x)$ and $V_g(x)$, iteratively.

As no existing method navigates from proprioception-only sparse contact measurements, the baselines are controlled ablations sharing \algoname's perception, mapping, and control stack, differing only in decision-making.

The Naive Goal Heuristic directly commands the legged robot toward the goal without considering terrain safety risks. The Safe-Goal Heuristic incorporates terrain safety estimation but does not explicitly reason about safe-region expansion.  The Pareto-Goal Heuristic iteratively selects frontier subgoals from the Pareto set defined by the goal value $V_{\mathrm{goal}}$ and the expansion probability $P_e$, thereby preserving the multi-objective structure without scalarization. To select a single subgoal from the Pareto set, PGH applies an \emph{a priori} rotating preference schedule: at each decision step it prioritizes one objective and breaks ties using the other. In our implementation, the schedule rotates in a repeating pattern of \{goal, goal, expansion\}, biasing selection toward goal progress while periodically forcing safe-set expansion.

We conduct experiments in two environments. Environment 1 (Fig.~\ref{fig:chrono}A) features a smooth transition in terrain penetration resistance, resulting in gradual changes in slip ratio and traversability. Environment 2 (Fig.~\ref{fig:chrono}B) features highly heterogeneous terrain characterized by abrupt spatial variations in penetration resistance, inducing sharp transitions in traversability.

In both environments, the robot is initialized within a verified safe region with three proprioceptive samples to bootstrap the GP terrain model. We evaluate three tasks: (1) goal-directed navigation in Environment 1, (2) goal-directed navigation in Environment 2, and (3) safe terrain exploration in Environment 2 without a predefined goal. All methods share identical GP hyperparameters, safety thresholds, and controller settings.

Performance is measured using: 
(1) success rate, defined as reaching the goal while maintaining full compliance with the ground-truth safety constraint; (2) completion time; and (3) total path length. Each experiment is executed three times per environment, and reported results are averaged across trials. 

\subsection{Experimental Results}

\begin{figure}[h]
    \centering
    \includegraphics[width=1\linewidth]{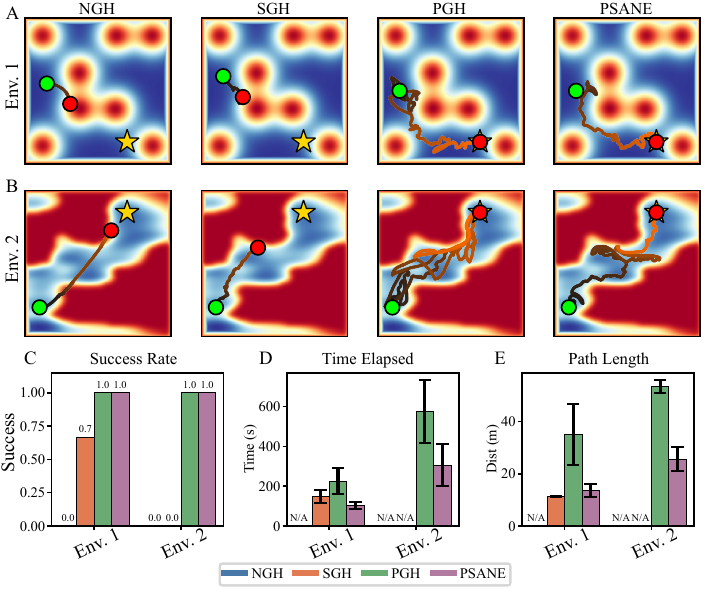}
    \caption{Navigation-to-goal comparison. (A,B) Representative paths in Environments 1 and 2. Green: start; red: final position; yellow: goal; blue--yellow--red: low-to-high ground-truth slip ratio. (C--E) Success rate, completion time, and path length averaged over three trials.}
    \label{fig:navigation}
\end{figure}

We evaluate \algoname\ against NGH, SGH, and PGH for goal-directed navigation in Environment~1 (Fig.~\ref{fig:navigation}A) and Environment~2 (Fig.~\ref{fig:navigation}B), and quantitative results are shown in Fig.~\ref{fig:navigation}C--E. For the safe exploration task in Environment~2, we compare \algoname\ against PGH to isolate the effect of scalarized frontier selection under safety constraints.
\subsubsection{Goal-Directed Navigation}
In both Environment~1 and Environment~2, NGH tends to follow nearly straight trajectories toward the goal. As a result, it enters regions with extremely high slip ratios, leading to immobilization and failure, resulting in a 0\% success rate in both environments. SGH incorporates safety estimation and reduces unsafe incursions. However, because it does not explicitly encourage safe-set expansion, it often stalls near the boundary of $S_t$, repeatedly selecting subgoals with high cost-to-go values but limited feasibility. When the unsafe region is narrow, the robot may eventually progress after prolonged stagnation because the subgoal is projected onto the interior of the safe boundary (Fig. \ref{fig:navigation}B, SGH); nevertheless, SGH fails to traverse large unsafe regions and frequently becomes trapped at frontiers (Fig. \ref{fig:navigation}A, SGH), yielding success rates of 66.7\% and 0\% in Environments~1 and~2.

In contrast, PGH and PSANE achieve 100\% success rates in both environments by explicitly considering both the cost-to-go value and safe-set expansion probability. The certified safe region guarantees trajectory safety, while the consideration of expansion probability enables the robot to enlarge the safe set and recover a feasible path whenever one exists. Beyond success rate, \algoname\ achieves more than a twofold improvement in efficiency, measured by reduced completion time (111.81~$\pm$~17.93\,s) and shorter path length (14.54~$\pm$~2.41\,m), compared to PGH (209.02~$\pm$~64.07\,s and 31.89~$\pm$~11.60\,m, respectively) in Environment 1. This gain arises from cost-aware expansion: unlike PGH, which alternates between expansion and goal-seeking phases and may detour at phase switches (Fig. \ref{fig:navigation}A, \ref{fig:navigation}B, PGH), \algoname\ jointly weighs expansion probability and cost-to-go through $V_{\mathrm{overall}}$, favoring lower-cost frontiers among similar expansion candidates and avoiding unnecessary detours (Fig. \ref{fig:navigation}A, \ref{fig:navigation}B, PSANE).

Moreover, in Environment~2, we introduce stronger spatial gradients and sharper local variations, increasing model uncertainty and making the task more challenging for all methods. The success of both PGH and \algoname\ in this setting demonstrates the effectiveness of the proposed safety characterization and indicates that it generalizes robustly under moderate environmental variability.
\begin{figure}[h]
    \centering
    \includegraphics[width=\linewidth]{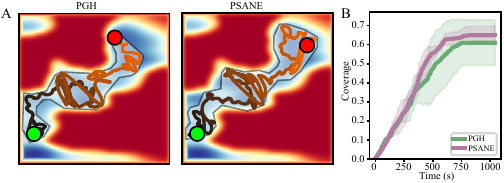}
    \caption{Comparison of exploration behavior and performance between PGH and \algoname. (A) Path and safe-region visualization. The black contour outlines the final certified safe region learned by the algorithm, with the green dot indicating the start position and the red dot the final position. (B) Coverage over time during the exploration phase. Coverage is computed as the ratio of the certified safe region area to the ground-truth safe region area.}
    \label{fig:exploration}
\end{figure}
\subsubsection{Safe Exploration Task}
We further demonstrate that the proposed framework generalizes to safe exploration tasks in which no predefined goal is specified. In this setting, the objective is to efficiently expand the certified safe region while maintaining conservative safety guarantees, as the robot autonomously collects interaction measurements from the environment. We conduct exploration simulations for 1000\,s in Environment~2.

Both PGH and PSANE safely explore the environment (Fig.~\ref{fig:exploration}A, black contour), progressively enlarging the certified safe region. When the explored area remains relatively small, PSANE achieves coverage growth comparable to PGH. However, as the explored region expands (e.g., after approximately 400\,s), PSANE demonstrates improved efficiency due to its cost-aware frontier selection, resulting in greater overall coverage (0.67~$\pm$~0.05) compared to PGH (0.59~$\pm$~0.12).

Notably, the lower region of the environment remains unexplored. Although this area does not violate the safety threshold, it exhibits relatively high predicted risk and low expansion score. Consequently, the robot prioritizes exploration of the upper region, where expansion potential and safety confidence are higher.

\subsection{Computational Performance}
\label{sec:ComputationalPerformance}
All components run online on a single 8-core AMD Ryzen~7 5800X3D CPU. The reactive controller runs at a fixed 10\,Hz, and the mapping and planning pipeline updates with each contact measurement. Per cycle on the $100 \times 100$ query grid ($10\,\text{m} \times 10\,\text{m}$ at $0.1$\,m resolution), GP inference takes 30\,ms (median), safe-set expansion and frontier extraction ${\sim}2$\,ms, and frontier scoring ${\sim}3$\,ms; the full callback completes in 62\,ms, well below the 0.8\,s measurement interval. Exact GP inference scales as $O(n^3)$ in the number of measurements, tractable here, while larger environments would require sparse or local GP approximations.

\section{Conclusion and Discussion}
\label{sec:Conclusion}
We presented \algoname, a proprioceptive safe active navigation and exploration framework for legged robots operating in unknown deformable granular environments where exteroceptive geometry is insufficient to reliably assess mobility risk. 
\algoname\ learns a GP-based slip-ratio model from sparse leg--terrain interaction measurements and uses uncertainty-aware confidence bounds to certify and iteratively expand a conservative safe region. A frontier-based subgoal planning algorithm integrated with a reactive controller enables simultaneous safe-set expansion and goal-directed navigation, while avoiding uncertified regions during execution. 
Simulations parameterized by field-measured LHS-1 terramechanics data show that \algoname\ improves success rate and maintains efficient trajectories relative to NGH, SGH, and PGH in both smoothly varying and highly heterogeneous environments.

While validation remains in simulation, the main sim-to-real gaps are identifiable: force--depth sensing noise propagates into resistance estimates, the slip model loses fidelity near regime boundaries, and gait variability perturbs contact geometry. These effects chiefly inflate measurement noise and model uncertainty, which the conservative certification absorbs as slower safe-set expansion, suggesting graceful rather than catastrophic degradation on hardware. The frontier objective is likewise an interpretable scalarized approximation of a planning problem intractable to solve exactly online; hardware validation, more principled subgoal selection (e.g., POMDP- or learning-based), and online estimation of the Lipschitz constant are important next steps.

This work establishes a framework for integrating interaction-derived terrain models into safety-critical navigation under uncertainty. By formalizing navigation on deformable terrain as a coupled estimation–exploration–control problem with explicit safety certification, PSANE provides a foundation for longer-horizon decision-making, multi-robot coordination, and multimodal terrain reasoning in future planetary exploration systems.

\smallskip
\noindent\textbf{Acknowledgments.} This work was supported by NASA LuSTR \#80NSSC24K0127, NASA PSTAR \#80NSSC22K1313, NASA MEP Technology Development Funding, NSF CAREER award \#2240075, and NSF FRR award \#2529696.

\bibliographystyle{IEEEtran}
\bibliography{references}

\end{document}